\documentclass[letterpaper, 10 pt, conference]{ieeeconf}  

\IEEEoverridecommandlockouts                              

\usepackage{lipsum}
\usepackage{xcolor}
\usepackage{amsmath}
\usepackage{amssymb}
\usepackage{bm}
\usepackage{graphicx}

\newcommand{\vect}[1]{\boldsymbol{\mathbf{#1}}}

\usepackage[style=ieee]{biblatex}

\DeclareSourcemap{
  \maps{
    \map{
      \pertype{article}
      \step[fieldset=language, null]
      \step[fieldset=url, null]
      \step[fieldset=doi, null]
      \step[fieldset=issn, null]
      \step[fieldset=isbn, null]
      \step[fieldset=note, null]
      \step[fieldset=editor, null]
      \step[fieldset=urldate, null]
      \step[fieldset=file, null]
    }
  }
}
\DeclareSourcemap{
  \maps{
    \map{
      \pertype{inproceedings}
      \step[fieldset=language, null]
      \step[fieldset=url, null]
      \step[fieldset=doi, null]
      \step[fieldset=issn, null]
      \step[fieldset=isbn, null]
      \step[fieldset=note, null]
      \step[fieldset=editor, null]
      \step[fieldset=urldate, null]
      \step[fieldset=file, null]
    }
  }
}
\DeclareSourcemap{
  \maps{
    \map{
      \pertype{incollection}
      \step[fieldset=language, null]
      \step[fieldset=url, null]
      \step[fieldset=doi, null]
      \step[fieldset=issn, null]
      \step[fieldset=isbn, null]
      \step[fieldset=note, null]
      \step[fieldset=editor, null]
      \step[fieldset=urldate, null]
      \step[fieldset=file, null]
    }
  }
}

\title{\LARGE \bf
Dynamic modeling of wing-assisted inclined running with a morphing multi-modal robot
}

\author{Eric Sihite$^{1}$, Alireza Ramezani$^{2*}$, and Morteza Gharib$^{1}$
\thanks{$^{1}$ Authors are with the Aerospace Engineering Department, California Institute of Technology, Pasadena, USA. Emails:
        {\tt\small esihite, mgharib@caltech.edu}}%
\thanks{$^{2}$ Authors are with the Silicon Synapse Lab, Department of Electrical and Computer Engineering, Northeastern University, Boston, USA. Emails:
        {\tt\small a.ramezani@northeastern.edu}}%
\thanks{$^{*}$ The corresponding author.}%
}

\begin{document}

\maketitle
\thispagestyle{empty}
\pagestyle{empty}

\begin{abstract}

Robot designs can take many inspirations from nature, where there are many examples of highly resilient and fault-tolerant locomotion strategies to navigate complex terrains by using multi-functional appendages. For example, Chukar and Hoatzin birds can repurpose their wings for quadrupedal walking and wing-assisted incline running (WAIR) to climb steep surfaces. We took inspiration from nature and designed a morphing robot with multi-functional thruster-wheel appendages that allows the robot to change its mode of locomotion by transforming into a rover, quad-rotor, mobile inverted pendulum (MIP), and other modes. In this work, we derive a dynamic model and formulate a nonlinear model predictive controller to perform WAIR to showcase the unique capabilities of our robot. We implemented the model and controller in a numerical simulation and experiments to show their feasibility and the capabilities of our transforming multi-modal robot.

\end{abstract}

\section{Introduction}







Robot designs can take many inspirations from nature, where there are many examples of highly resilient and fault-tolerant locomotion strategies to navigate complex terrains by using multi-functional appendages. For instance, amphibious animals such as sea lions utilize their limbs for swimming and terrestrial walking \cite{kerr2022biomechanical}, or birds such as Chukars and Hoatzins can repurpose wings for quadrupedal walking and wing-assisted incline running (WAIR) \cite{tobalske2007aerodynamics, dial2003wing}. These animals showcase impressive dexterity in employing the same appendages in different ways and achieve multiple modes of locomotion, resulting in highly plastic locomotion traits that enable them to interact and navigate various environments and expand their habitat range. 

\begin{figure}[t]
\vspace{0.08in}
    \centering
    \includegraphics[width=0.9\linewidth]{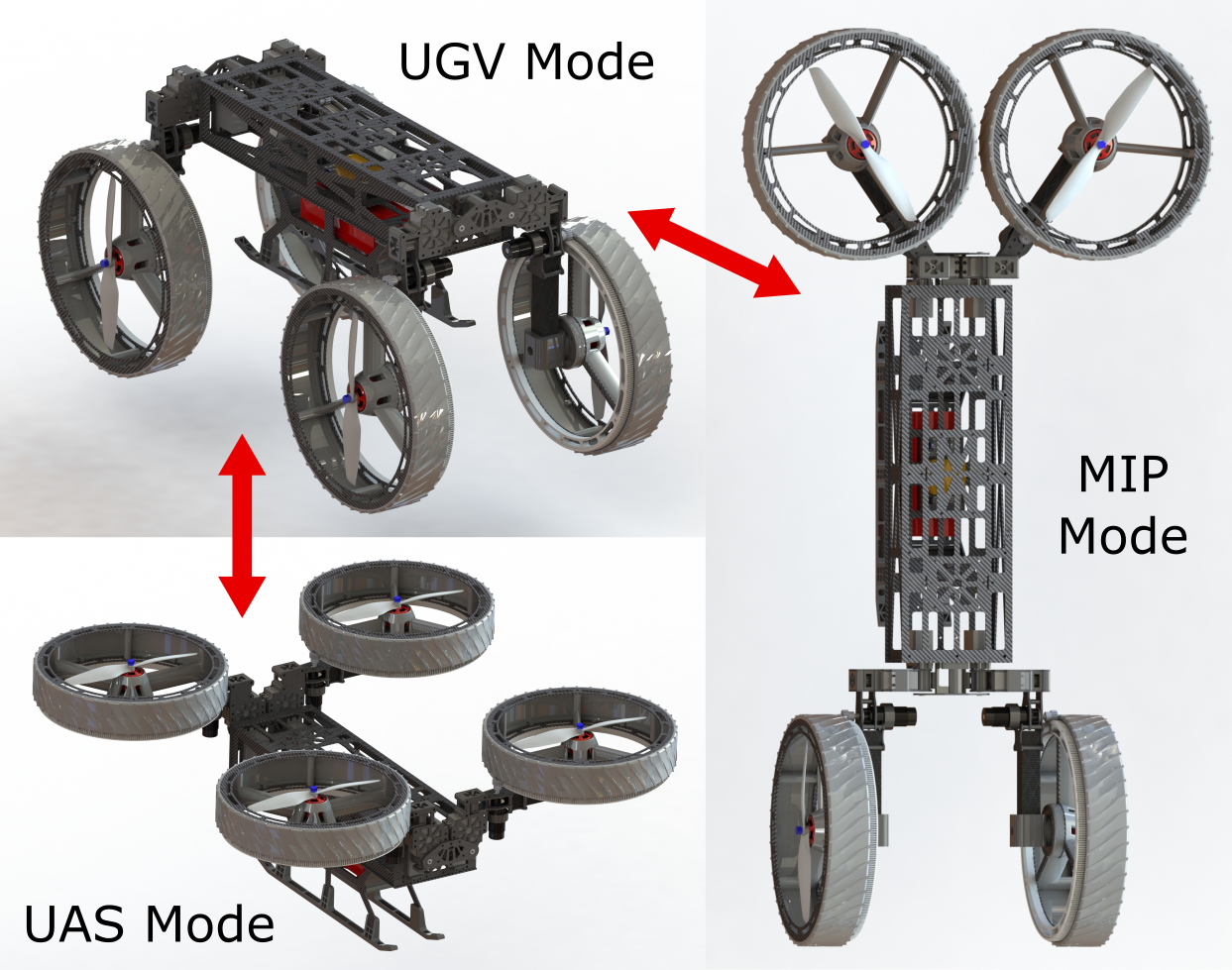}
    \caption{Shows the illustrations of our multi-modal morphing robot that is capable of transforming between the unmanned ground vehicle (UGV), unmanned aerial system (UAS), and mobile inverted pendulum (MIP) configurations.}
    \label{fig:robot_overview}
\vspace{-0.08in}
\end{figure}

The robotic biomimicry of animals' appendage repurposing can yield mobile robots with unparalleled capabilities. Taking inspiration from animals, we have designed a robot capable of negotiating unstructured, multi-substrate environments, including land and air, by employing its components in different ways as wheels, thrusters, and legs. This robot, shown in Fig.~\ref{fig:robot_overview}, can employ its multi-functional components composed of several actuator types to (1) fly, (2) roll, (3) crawl, (4) crouch, (5) balance, (6) tumble, (7) scout, and (8) loco-manipulate. Our design is unique compared to other multi-modal robots, such as soft robots which utilize body morphing for mobility \cite{ishida2019morphing, shah2021soft, joyee20223d, kotikian2019untethered}, robots with redundant actuators that simply combines two separate designs into one single robot (e.g., quad-rotor with wheels \cite{araki2017multi}, or wheeled-legged robots \cite{bjelonic2019keep, grand2004stability, ijspeert2007swimming, riviere2018agile}). On the other hand, our robot aims to utilize body morphing and appendage repurposing to achieve multi-modality by using its multi-purpose wheel-thruster appendages for actuation. By doing so, we seek to achieve the versatility and fault tolerance that is seen in animals in nature.

\begin{figure}[t]
\vspace{0.08in}
    \centering
    \includegraphics[width=\linewidth]{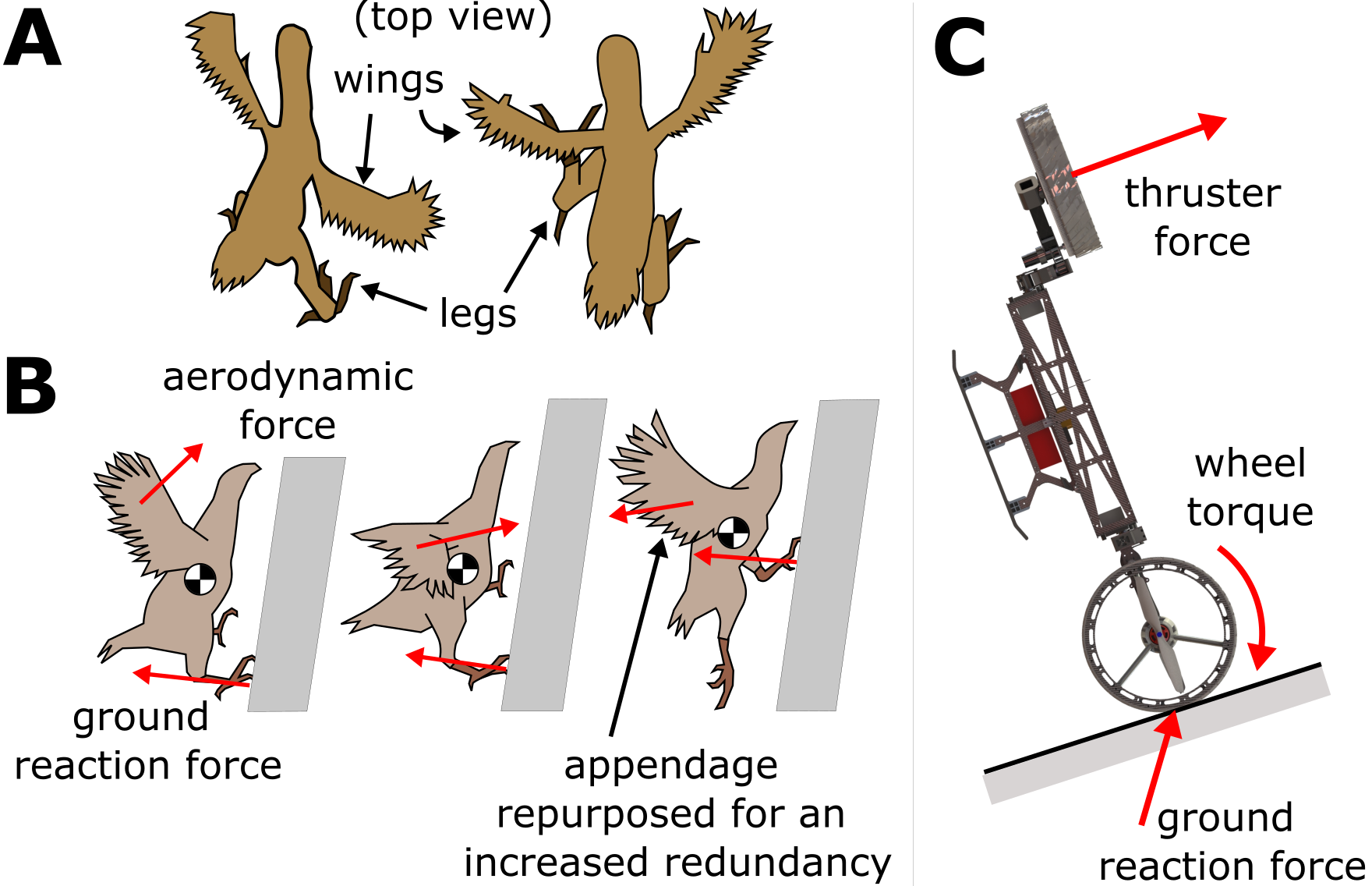}
    \caption{Illustrate the cartoon depiction of appendage repurposing. \textbf{(A)} Hoatzin nestlings wing-assisted quadrupedal locomotion \cite{abourachid2019hoatzin}. \textbf{(B)} Chukar birds' wing-assisted incline running (WAIR) \cite{dial2003wing}. \textbf{(C)} Our robot utilizes its thrusters to perform WAIR on an inclined surface.}
    \label{fig:wair}
\vspace{-0.08in}
\end{figure}

In this work, we utilize our transforming robot, the multi-modal mobility morphobot (M4) \cite{sihite2023multi}, which is capable of switching across several mobility modes including ground, aerial, and mobile inverted pendulum (MIP) modes. In this work, we attempt to utilize our robot's MIP locomotion mode and use its thrusters to perform WAIR and assist in climbing an inclined surface, as illustrated in Fig.~\ref{fig:wair}. The addition of thrusters in our robot makes the control design unique compared to the other conventional MIP robots, such as \cite{lee2008control, sihite2018attitude, rigatos2020nonlinear, varghese2017optimal}, and we must enforce the ground contact forces constraint to prevent wheel slip on the inclined surface. The enforcement of the ground constraints can be achieved by using an optimization-based controller, such as model predictive control (MPC). The framework required for this is derived from control of legged robots, which are often modeled as a simple inverted pendulum \cite{ramezani2014performance, grizzle2013progress}. It draws from previous work on ground contact regulation for thruster assisted locomotion in legged robots \cite{liang2021rough, sihite2021unilateral, dangol2021control, dangol2020performance, de2020thruster}. However, this work is distinct as wheeled locomotion does not benefit from foot placement abilities of legged robots, adding more constraints to the model. Our novel contributions in this paper are the concept design of a unique morphing multi-modal robot, the derivation of a nonlinear model of a thruster-assisted MIP on an inclined surface, and the implementation of a nonlinear MPC of our robot performing WAIR in a simulation and experiments.

This paper is outlined as follows: we begin with the introduction which is followed by a brief overview of the robot's mechanical design and capabilities, dynamic modeling of the unreduced model and the reduced-order thruster-assisted MIP model, controller design, and MPC design formulations. Finally, we present the simulation results, discussions, and concluding remarks.

\section{Brief Overview of the Robot}

Our transforming robot, shown in Fig.~\ref{fig:robot_overview}, can switch its modes of mobility between unmanned ground vehicle (UGV), unmanned aerial system (UAS), mobile inverted pendulum (MIP), quadrupedal, thruster-assisted MIP, legged locomotion, and manipulation. The robot possesses an articulated body with four legs, each leg has a total of two hip degrees of freedom (DOF) and a shrouded propeller that act as a wheel and a thruster simultaneously. Eight independent joint actuators are employed to translate the legs forward-backward and sideways, and the shrouded propellers are attached to the leg ends. Among the modes listed above, the work in this paper focuses on the transformation from UGV to MIP configuration and performing WAIR using the thrusters attached to the upper legs as the robot drives on an inclined surface. 

The robot weighs approximately 6.0 kg with all components, which include the onboard computers for low-level control and data collection, sensors (encoders, inertial measurement unit, stereo cameras), communication devices for teleoperation, and power electronic components. The majority of the robot's weight belongs to the power components consisting of 4 wheel motors, 4 propeller motors, 8 joint servos, motor drivers, and a 6S 4000mAh battery. 

The robot measures 0.7 m in length, and 0.35 m in both width and height when in UGV mode. When in the MIP mode and dynamically balancing on its two wheels, it is 1.0 m tall, which permits reaching a better vantage point for data collection using its exteroceptive sensors. When in UAS configuration, the robot is 0.3 m tall, and the propellers' center points can reach a maximum distance of 0.45 m far apart from each other. Each propeller-motor combination can generate a maximum thrust force of approximately 2.2 kg-force, therefore reaching roughly 9 kg thrust force in total for an approximately 1.5 thrust-to-weight ratio. Its legs including the wheels are 0.3 m long, and its wheels are 0.25 m in diameter, which allows for traversing bumpy terrain.

\section{Dynamic Modeling}

The dynamic modeling for the simulation and controller model can be derived using Euler-Lagrangian formulations by deriving conservative energy equations and mapping the non-conservative forces into the generalized coordinates \cite{landau1960mechanics}. Two models are derived in this section, where one is the unreduced model that is used in the simulation and the other is the 2D MIP reduced order model for designing the controller and MPC.



\subsection{Unreduced robot model for simulation}

\begin{figure}[t]
\vspace{0.08in}
    \centering
    \includegraphics[width=\linewidth]{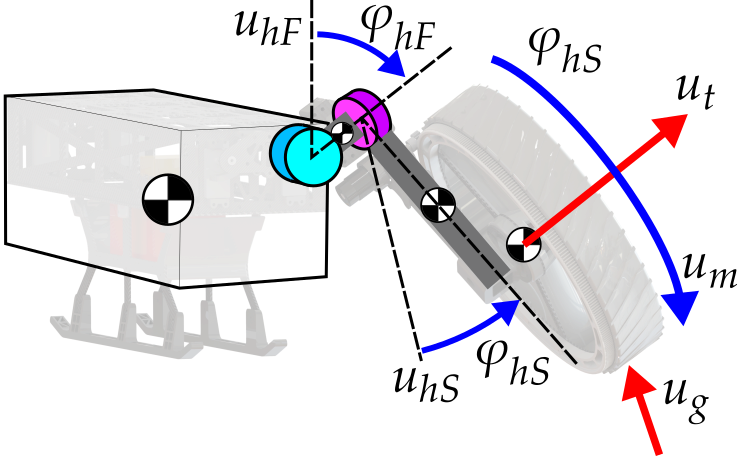}
    \caption{Mechanical configuration of the robot's appendages, showing the degrees of freedom and actuation of one of its legs. $\varphi_{hF}$, $\varphi_{hS}$, and $\varphi_{w}$ are the hip frontal, sagittal, and wheel angles, respectively. $u_{hF}$, $u_{hS}$, $u_m$, and $u_t$ are the hip frontal, hip sagittal, motor, and thruster actuation, respectively. Additionally, $\vect u_g$ is the ground contact force if the wheels are in contact with the ground.}
    \label{fig:m4_dof}
    \vspace{-0.08in}
\end{figure}

The robot is composed of thirteen rigid bodies consisting of one main body and four appendages, where each appendage is composed of two linkages and one shrouded propeller (wheel), as illustrated in Fig.~\ref{fig:m4_dof}. For each $j$-th massed body, $m_j$ denotes the mass, $\hat{\vect{I}}_j \in \mathbb{R}^{3 \times 3}$ is the principal inertia tensor, $\vect p_j \in \mathbb{R}^{3}$ and $\vect v_j \in \mathbb{R}^{3}$ are the inertial position and velocity vectors, respectively, and $\vect{\omega}_j \in \mathbb{R}^{3}$ is the angular velocity vector defined in the body frame. Furthermore, let $g$ be the gravitational acceleration and $\vect g = [0,0,-g]^\top$ be the acceleration vector defined in the inertial frame. Then, the Lagrangian $\mathcal{L}$ can be written as a function of the conservative energy in the system:
\begin{equation}
    \mathcal{L} = \sum_{j} \frac{1}{2} \left( m_j \vect v_j^\top \vect v_j  + \boldsymbol{\omega}_j^\top \hat{\vect{I}}_j  \boldsymbol{\omega}_j \right) + m_j \vect p_j^\top \vect g,
\label{eq:energy}
\end{equation}
where the first two terms are the linear and angular kinetic energy, while the last term is the potential energy. 

Let $\vect q \in \mathbb{R}^{18}$ be the generalized coordinates of the system, which consist of the body's 6 degrees of freedom (DOF) (position and orientation) and the legs' 12 DOF. Then, the equation of motion can be derived as follows:
\begin{equation}
    \frac{d}{dt} \left( \frac{ \partial \mathcal{L} }{ \partial \dot{\vect q} } \right) - \frac{ \partial \mathcal{L} }{ \partial\vect q } =  \sum_i \vect{B}_k \, \vect u_k,
\label{eq:eom_full}
\end{equation}
where $\vect u_i$ are the combined non-conservative forces and torques acting on leg $i \in \{1, \dots, 4\}$, and $\vect B_i$ is the mapping of $\vect u_i$ to the generalized coordinates $\vect q$. The mapping $\vect B_i$ can be derived from the principle of virtual work while the ground contact forces $\vect u_{g,i}$ can be modeled using the Stribeck friction model and the compliant ground model \cite{armstrong1990stick}. The compliant ground model simply models the ground as spring and damper with large stiffness and damping coefficients to calculate the normal forces. Then, the friction forces can be calculated as a function of the normal forces and sliding speed of the wheel's contact point relative to the ground. 

Finally, \eqref{eq:eom_full} can be derived into the following general form:
\begin{equation}
\dot{\vect x} = \vect{f}(\vect x, \vect u),
    \label{eq:eom_full_ss}
\end{equation}
where $\vect x = [\vect q^\top, \dot{\vect q}^\top]^\top$ is the system states and $\vect u$ is the combined form of all inputs $\vect u_i$. This equation of motion is highly nonlinear and it will be difficult to design a controller using this model. Therefore, the unreduced model in \eqref{eq:eom_full_ss} was used to simulate the robot in our simulator in Section~\ref{sec:simulation} while the controller design used a reduced order model that will be derived in the following section.

\subsection{Reduced order model (2D MIP) for control design}
\label{subsec:2d_mip_dynamics}

\begin{figure}[t]
\vspace{0.08in}
    \centering
    \includegraphics[width=\linewidth]{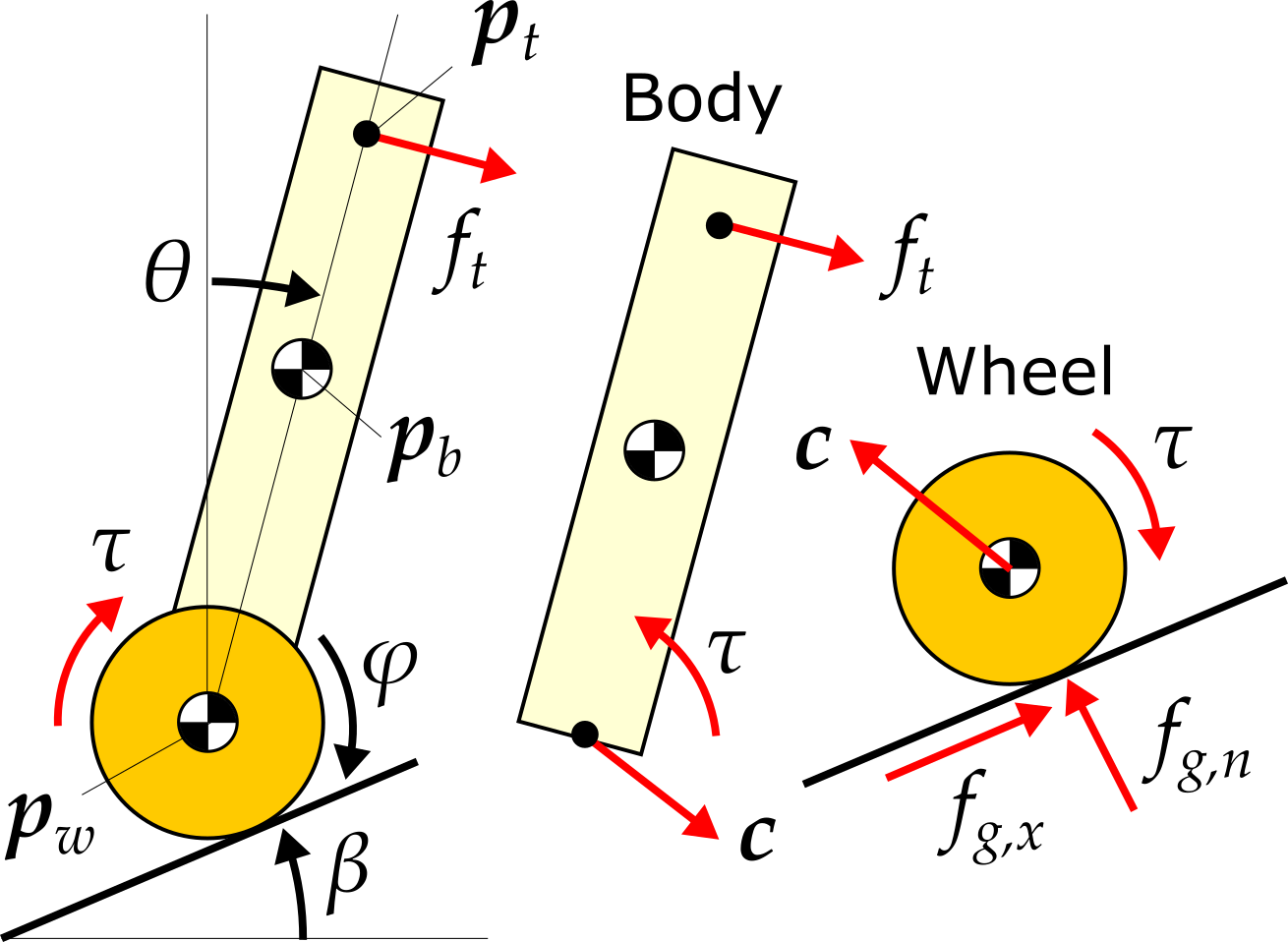}
    \caption{Free-body diagram (FBD) of the 2D thruster-assisted mobile inverted pendulum system. The system consists of two bodies (both wheels are combined into a single body).}
    \label{fig:fbd_2d}
    \vspace{-0.08in}
\end{figure}

A 2D thruster-assisted mobile inverted pendulum dynamic equation of motion is derived for the controller design. Figure \ref{fig:fbd_2d} shows the free-body diagram of the system with states, forces, and torques acting on the system. Let $\vect{q} = [\phi, \theta]^\top$ be the generalized coordinates of the reduced order model, where $\phi$ and $\theta$ are the wheel and body pitch angles, respectively. Assuming that the wheels are not slipping with respect to the ground, the wheel, body center of mass (CoM), and thruster positions ($\vect{p}_w$, $\vect{p}_b$, and $\vect{p}_t$, respectively. $\vect{p}_w,\vect{p}_b,\vect{p}_t \in \mathbb{R}^2$) can be derived as follows:
\begin{equation}
\begin{aligned}
    \vect{p}_w &= \phi \, r 
    \begin{bmatrix} \cos (\beta) \\ \sin (\beta) \end{bmatrix}, \quad &
    \begin{aligned}
      \vect{p}_b &= \vect{p}_w + \vect{R}(\theta)\, \vect{\ell}_b,\\
      \vect{p}_t &= \vect{p}_w + \vect{R}(\theta)\, \vect{\ell}_t,
    \end{aligned} 
\end{aligned} 
\label{eq:pos_vectors}
\end{equation}
where $\vect{R}(\theta) \in SO(2)$ is the 2D rotation matrix, $r$ is the wheel radius, $\beta$ is the floor inclination angle, $\vect{\ell}_b$ and $\vect{\ell}_t$ are constant body-frame length vectors from the wheel center of rotation (CoR) to the body CoM and thrusters, respectively. The wheel motor torques ($\tau$) and thruster forces ($f_t$) are the external forces of the system, which can be derived using the principle of virtual work. The mapping into the generalized coordinates can be derived as follows:
\begin{equation}
\begin{aligned}
    \vect{B}_\tau &= \left( \frac{\partial (\dot \phi - \dot \theta)}{\partial \dot{\vect{q}}} \right)^\top, &
    \vect{B}_t &= \left( \frac{\partial \dot{\vect{p}}_t }{\partial \dot{\vect{q}}} \right)^\top 
    \begin{bmatrix}
        \cos(\theta) \\ -\sin(\theta)
    \end{bmatrix},
\end{aligned} 
\label{eq:virtual_work}
\end{equation}  
where $\vect{B}_\tau$ and $\vect{B}_t$ are the generalized force mapping for $\tau$ and $f_t$, respectively.

Then, the Euler-Lagrangian dynamics equation of motion can be derived as follows:
\begin{equation}
\begin{aligned}
    \mathcal{L} =& \, m_b \, \dot{\vect{p}}_b^\top \dot{\vect{p}}_b/2 + I_b \, \dot{\theta}^2/2 + m_b \, \vect{p}_b^\top \vect{g} \\
    &\, + m_w \, \dot{\vect{p}}_w^\top \dot{\vect{p}}_w/2 + I_w \, \dot{\phi}^2/2 + m_w \, \vect{p}_w^\top \vect{g},
\end{aligned} 
\label{eq:lagrangian_2d}
\end{equation}    
\begin{equation}
\begin{aligned}
    \frac{d}{dt}\left( \frac{\partial \mathcal{L}}{\partial \dot{\vect{q}}} \right) - \frac{\partial \mathcal{L}}{\partial \vect{q}} &= \vect{B}_\tau \, \tau + \vect{B}_t \, f_t,
\end{aligned} 
\label{eq:eom_2d}
\end{equation}    
where $\vect{g} = [0,-g]^\top$ and $g$ is the gravitational acceleration constant, $m_b$ and $m_w$ are the body and wheel masses, respectively, $I_b$ and $I_w$ are the body and wheel pitch moment of inertia, respectively. The electric motor torque can be modeled as:
\begin{equation}
\begin{aligned}
    \tau = k_1 \, u_m - k_2 \, \dot{\phi},
\end{aligned} 
\label{eq:motor_model}
\end{equation}    
where $k_1 > 0$ and $k_2 > 0$ are the motor constants, and $u_m \in [-1,1]$ is the motor controller PWM duty. Combining \eqref{eq:eom_2d} and \eqref{eq:motor_model} forms the following equation of motion: 
\begin{equation}    
\begin{aligned}
    \vect{M}(\vect{q})\, \ddot{\vect{q}} + \vect{h}(\vect{q}, \dot{\vect{q}}) = \vect{B}_q(\vect{q}) \, \vect{u},
\end{aligned} 
\label{eq:eom_2d_v2}
\end{equation}    
where $\vect{u} = [u_m, f_t]^\top$ is the input vector. Then, the equation of motion can be derived into the following state-space form:
\begin{equation}
\begin{aligned}
    \dot{\vect{x}} &= \underbrace{\begin{bmatrix}
        \dot{\vect{q}} \\ -\vect{M}^{-1}\,\vect{h}
    \end{bmatrix}}_{\vect{f}(\vect{x})} + \underbrace{\begin{bmatrix}
        \vect{0}_{2 \times 2} \\ \vect{M}^{-1}\,\vect{B}_q 
    \end{bmatrix}}_{\vect{g}(\vect{x})} \, \vect{u},
\end{aligned} 
\label{eq:nonlinear_model_2d}
\end{equation}    
where $\vect{x} = [\vect{q}^\top, \dot{\vect{q}}^\top]^\top$ is the state vector.



\subsection{Ground contact forces estimation}

\begin{figure*}[t]
\vspace{0.08in}
    \centering
    \includegraphics[width=\linewidth]{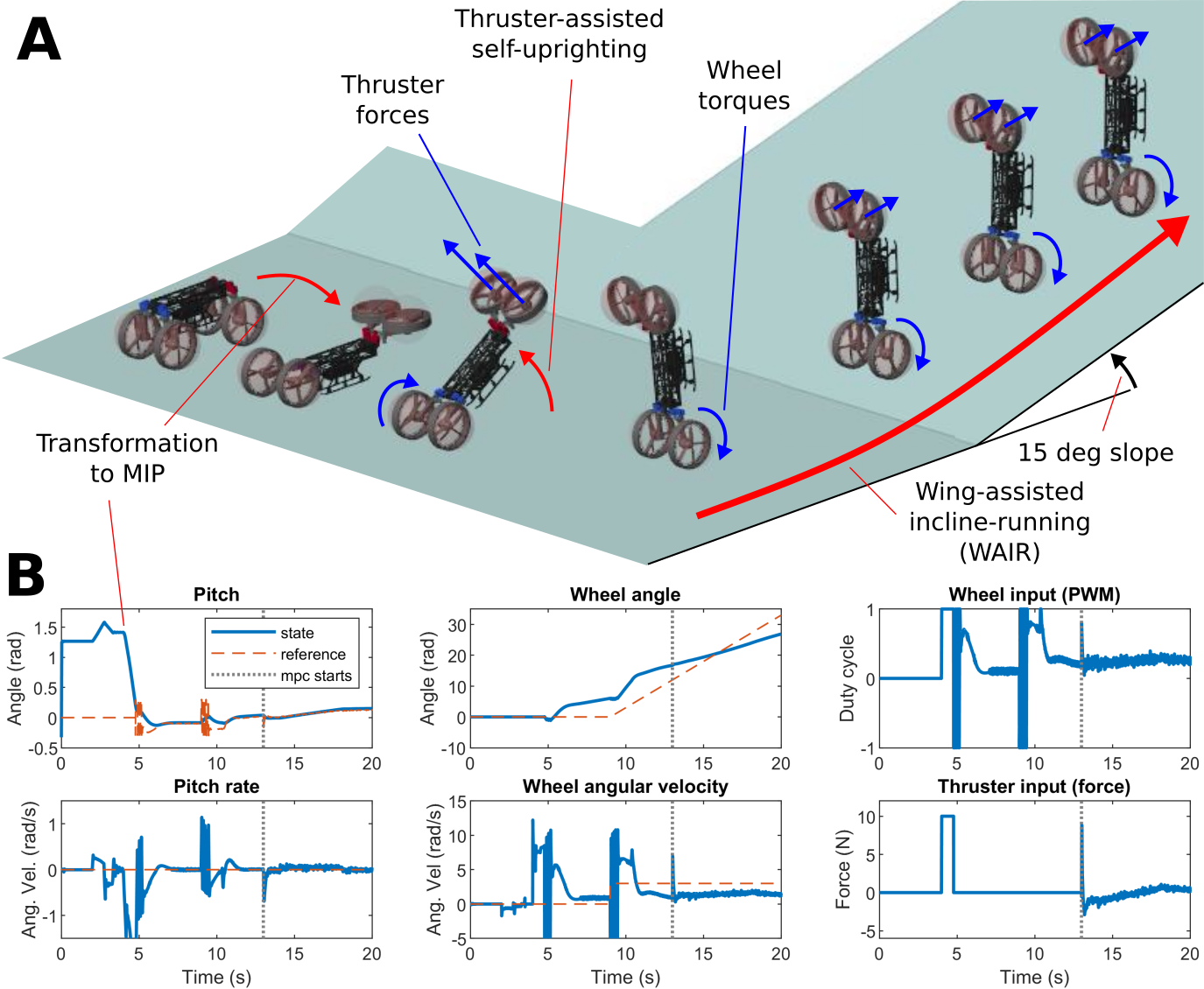}
    \caption{\textbf{(A)} Composite image and \textbf{(B)} plots of the simulation that shows the robot transforming from UGV into upright MIP configuration, then performing WAIR on an inclined surface. The robot was running on PID controller before the MPC was enabled at $t = 13$s when the robot was on the inclined surface.}
    \label{fig:plot_states}
    \vspace{-0.08in}
\end{figure*}

The model derived in Section~\ref{subsec:2d_mip_dynamics} assumes that the wheels do not slip with respect to the ground. Therefore, we must make sure that the ground contact forces follow the friction cone constraint. In order to do so, we must derive the ground contact forces as a function of both the states ($\vect x$) and inputs ($\vect u$). The free-body diagram in Fig.~\ref{fig:fbd_2d} can be used to derive Newton's second law of motion, as follows: 
\begin{align}
    m_w \, \ddot{\vect{p}}_w &= m_w \, \ddot \phi \, r 
    \begin{bmatrix} \cos (\beta) \\ \sin (\beta) \end{bmatrix} = \vect c + \vect{R}(-\beta) \, \begin{bmatrix}
        f_{g,x} \\ f_{g,n}
    \end{bmatrix}
    \label{eq:fbd_equations_1}\\
    I_w \, \ddot{\phi} &= \tau - r \, f_{g,x}
    \label{eq:fbd_equations_2}\\
    I_b \, \ddot{\theta} &= -\tau - \left( \vect{p}_w - \vect{p}_b \right) \times \vect{c} + \ell_{t,z}\,  f_t,
    \label{eq:fbd_equations_3}
\end{align}
where $\vect{c}$ is the contact force between the body and the wheel, $f_{g,x}$ and $f_{g,n}$ are the ground friction and normal forces, respectively, $l_{t,z}$ is the moment arm to the thruster force, and $\times$ in \eqref{eq:fbd_equations_3} is the 2D cross product. The acceleration terms in \eqref{eq:fbd_equations_1} to \eqref{eq:fbd_equations_3} can be derived from \eqref{eq:eom_2d_v2} by solving for $\ddot{\vect{q}}(\vect{x}, \vect{u})$. Let $\vect{z} = [\vect{c}^\top, f_{g,x}, f_{g,n}]^\top$ be the unknown variables in \eqref{eq:fbd_equations_1} to \eqref{eq:fbd_equations_3}. $\vect{z}$ can then be solved by reformulating \eqref{eq:fbd_equations_1} to \eqref{eq:fbd_equations_3} into  $\vect{A}_c(\vect{x}) \, \vect{z} = \vect{b}_c(\vect{x}, \vect{u})$. Then, the ground contact forces can be solved as follows:
\begin{equation}
\begin{aligned}
    \begin{bmatrix}
        f_{g,x} \\ f_{g,n}
    \end{bmatrix} &= \begin{bmatrix}
        \vect{0}_{2 \times 2}, & \vect{I}_{2 \times 2}
    \end{bmatrix}
    \vect{A}_c(\vect{x})^{-1} \, \vect{b}_c(\vect{x}, \vect{u}),
\end{aligned} 
\label{eq:ground_friction}
\end{equation}    
where $\vect{0}_{2 \times 2}$ and $\vect{I}_{2 \times 2}$ are empty and identity matrices, respectively. Note that $\vect{A}_c$ is singular if $\theta = \beta$. In this case, we use a workaround by averaging the $\vect z$ solved using $\theta + \Delta \theta$ and $\theta - \Delta \theta$ where $\Delta \theta$ is a small deviation from $\theta = \beta$. Finally, define the no-slip conditions between the wheels and the ground as follows:
\begin{equation}
\begin{aligned}
    f_{g,n} &> 0, \quad & (\mu_s \, f_{g,n})^2 & > f_{g,f}^2,
\end{aligned} 
\label{eq:ground_constraint}
\end{equation}   
where $\mu_s$ is the ground static friction coefficient. Both constraints in \eqref{eq:ground_constraint} can be reformulated into the simple form of $\vect{h}_c(\vect{x}, \vect{u}) < \vect{0}$ as the constraints in the MPC formulation.

\section{Controller and MPC Formulations}
\label{sec:control}

The robot can transform from the UGV configuration into the upright MIP position using joint actuation and thrusters, as illustrated in Fig.~\ref{fig:m4_experiment} and Fig.~\ref{fig:plot_states}-A. This was done in the following sequence: transform from the rover into the MIP configuration while sitting on the landing gears, and activate the thruster to propel the robot into the upright configuration. Once the robot is at the upright position, then the MIP stabilizing controller is activated to stabilize the robot about its unstable equilibrium. The landing sequence and transformation back into the UGV mode simply follow the same process in reverse.

If the robot operates on a flat surface, only the wheels are used to stabilize the robot and the controller follows a typical two-wheeled inverted pendulum stabilizing PID controller \cite{lee2008control}. As the robot climbs on an inclined surface, the wheel motors and thrusters must provide additional torques and forces to perform WAIR. The additional wrenches from this compensation can potentially cause the wheels to slip and violate the assumption used in \eqref{eq:nonlinear_model_2d}. Therefore, the MPC framework is used to determine the controller inputs so the robot can climb the inclined surface without violating the no-slip condition defined in \eqref{eq:ground_constraint}. 

We must first discretize the continuous model in \eqref{eq:nonlinear_model_2d} using a simple explicit Euler integration scheme:
\begin{equation}
\begin{aligned}
    \vect{x}_{k+1} &= \vect{x}_{k} + \Delta t \, \left( \vect f (\vect{x}_k) + \vect g (\vect{x}_k)\, \vect{u}_k \right)
\end{aligned} 
\label{eq:discrete_model}
\end{equation}   
where $\Delta t$ is the controller time step, $\vect{x}_{k} = \vect{x}(t_k)$, $\vect{u}_{k} = \vect{u}(t_k)$, and $t_k$ is the discrete time at discrete step $k$. Let $\vect{x}_{r,k}$ be the controller state reference and $\vect{e}_k = \vect{x}_{r,k} - \vect{x}_{k}$ be the state tracking error. Then, the MPC is set up as follows:
\begin{equation}
\begin{aligned}
    \min_{\vect{u}_k} \quad &
    J = \sum_{k=1}^{n_h} \left( \vect{e}_k^\top \,\vect{Q}\, \vect{e}_k + \vect{u}_{k-1}^\top \,\vect{R}\, \vect{u}_{k-1} \right) \\
    \textrm{s.t.} \quad &
    \vect{x}_{k+1} = \vect{x}_{k} + \Delta t \, \left( \vect f (\vect{x}_k) + \vect g (\vect{x}_k)\, \vect{u}_k \right) \\
    & \vect{h}_c(\vect{x}_{k}, \vect{u}_k) < \vect{0}, \quad \textrm{ for } k \in \{0,\dots, n_h-1\}\\
    &  \vect{u}_{min} \leq \vect{u}_k \leq \vect{u}_{max}
\end{aligned} 
\label{eq:mpc_formulation}
\end{equation}   
where $n_h$ is the MPC prediction horizon, $\vect{Q} \in \mathbb{R}^{4 \times 4}$ and $\vect{R} \in \mathbb{R}^{2 \times 2}$ are diagonal weighting matrices, $\vect{u}_{min}$ and $\vect{u}_{max}$ are the lower and upper bounds of $\vect{u}_{k}$, respectively. Note that the constraint $\vect{h}_c(\vect{x}_{k}, \vect{u}_k) < \vect{0}$ is applied for $k \in \{0,\dots, n_h-1\}$, resulting in a total of $2 n_h$ constraints.

\section{Results and Discussions}
\label{sec:simulation}


\begin{figure}[t]
\vspace{0.08in}
    \centering
    \includegraphics[width=\linewidth]{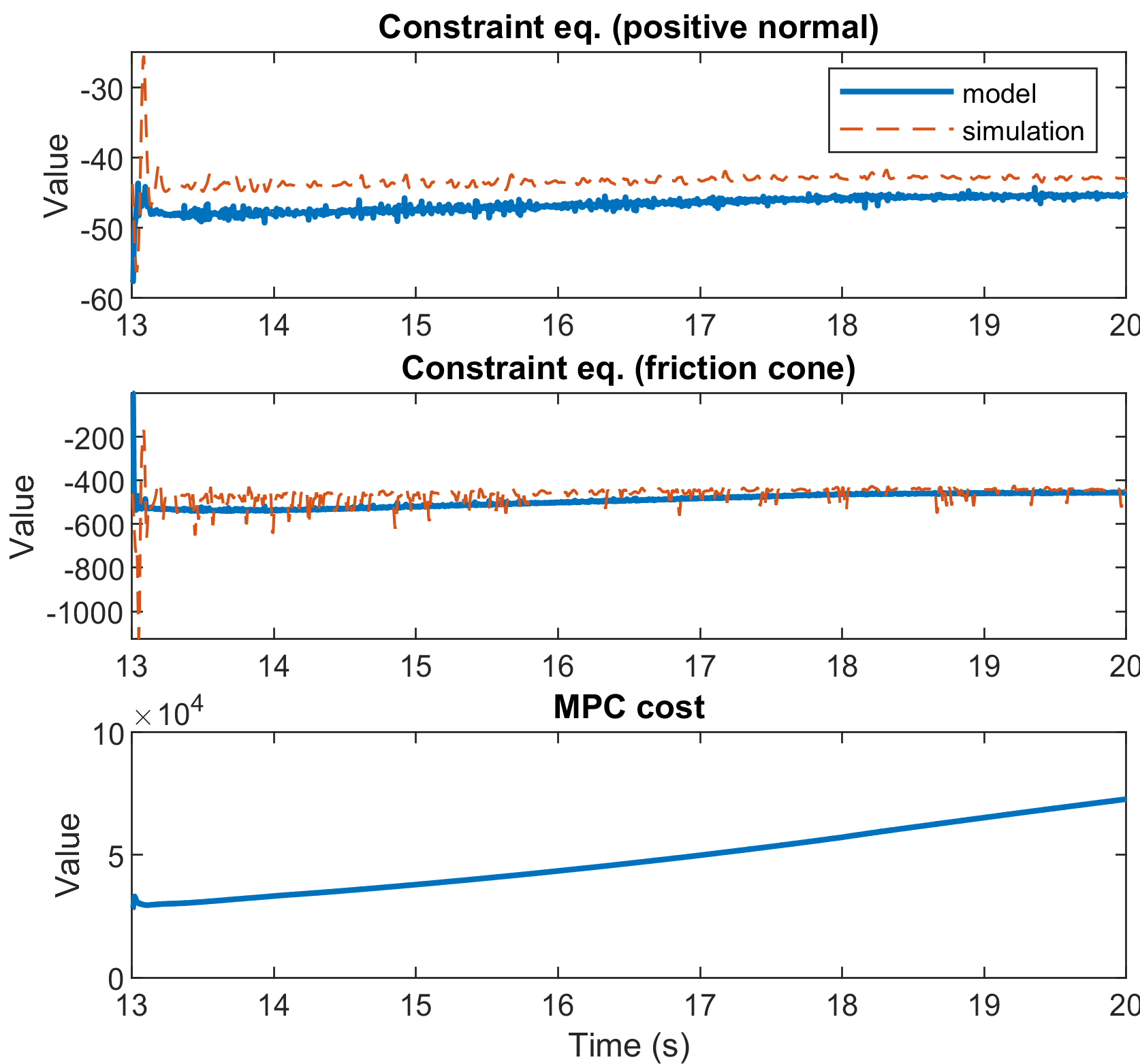}
    \caption{MPC constraint equations ($\vect{h}(\vect{x}_{k}, \vect{u}_k) < \vect{0}$) comparison between using the ground force model in \eqref{eq:ground_friction} and the simulator's ground force values, in addition to the MPC cost over time. Both constraint equations are similar between the model and the actual forces used in the simulator which shows the accuracy of the model. However, the cost is slowly increasing which is likely caused by the tracking error of the wheel position and velocities.}
    \label{fig:plot_mpc}
    \vspace{-0.08in}
\end{figure}

\begin{figure}[t]
\vspace{0.08in}
    \centering
    \includegraphics[width=\linewidth]{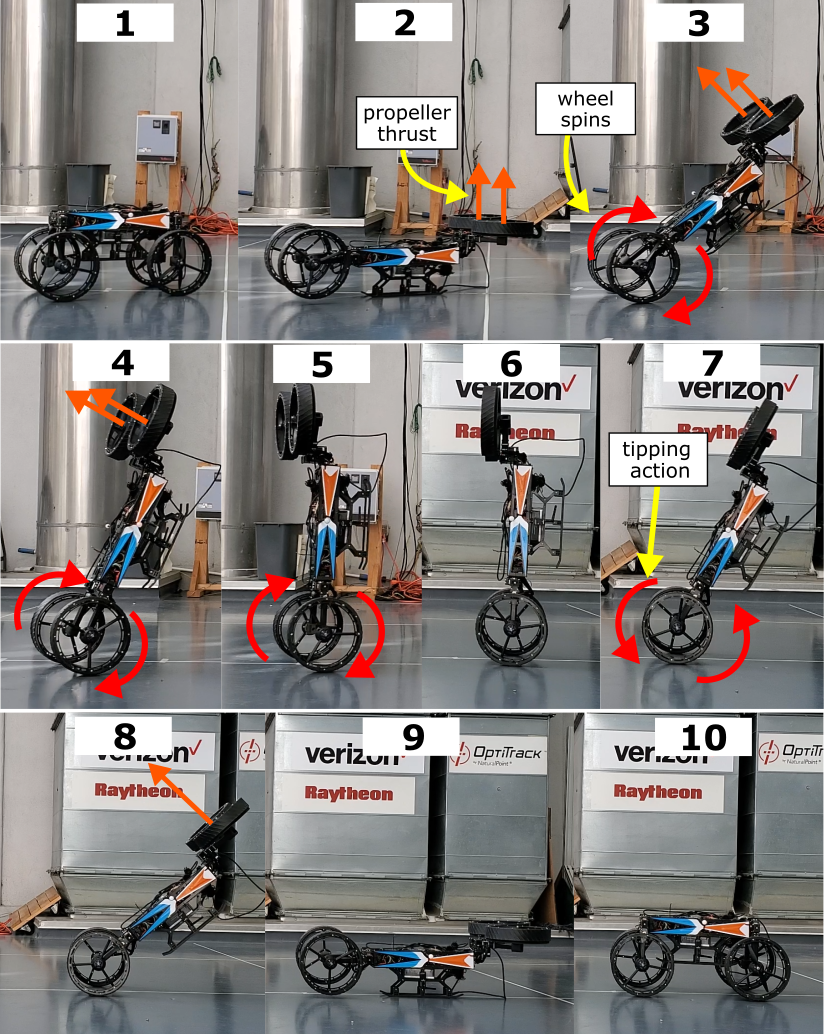}
    \caption{Time-lapsed images of the robot performing transformation from the UGV into the MIP configuration (1-2), thruster-assisted self-uprighting (3-5), MIP balancing (6), touch-down sequence and transformation back into UGV configuration (7-10).}
    \label{fig:m4_experiment}
    \vspace{-0.08in}
\end{figure}

\begin{figure}[t]
\vspace{0.08in}
    \centering
    \includegraphics[width=\linewidth]{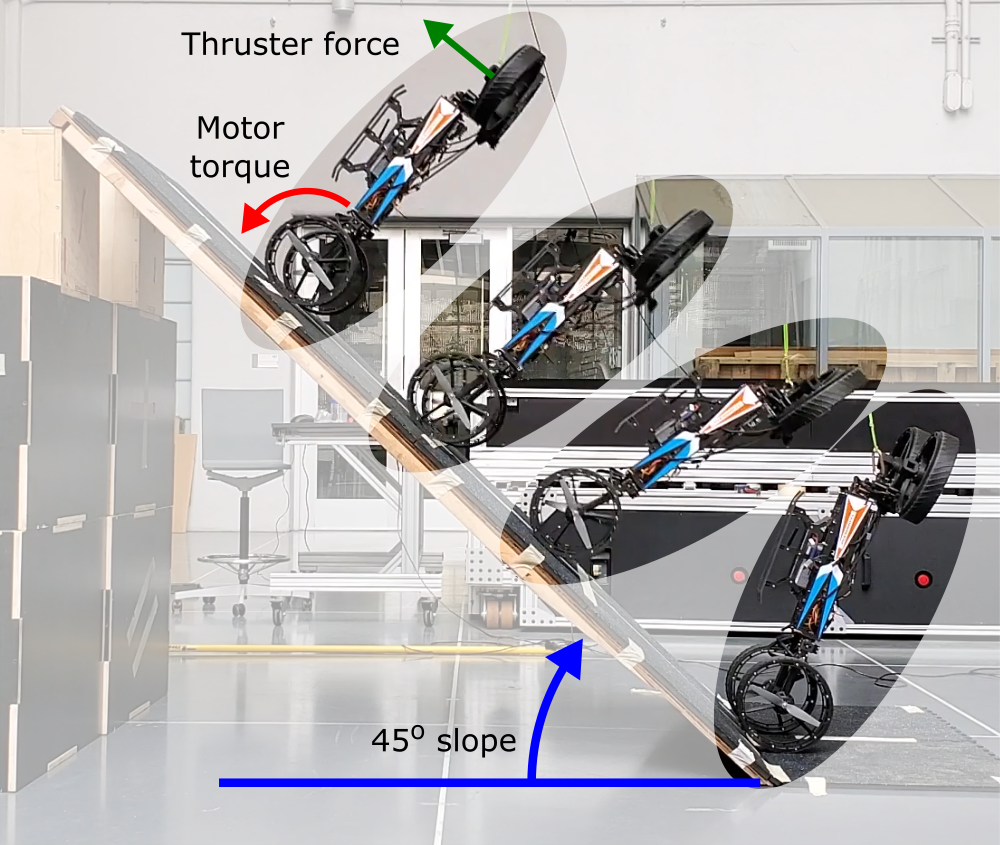}
    \caption{Time-lapsed image of the robot performing WAIR using MIP configuration with the help of the thrusters to climb a 45 degrees incline.}
    \label{fig:wair_experiment}
    \vspace{-0.08in}
\end{figure}

The simulation was performed using Simulink's Simscape multi-body and contact forces library, where we imported the robot's CAD model into the simulator with the appropriate mass and inertia properties. We then implemented the controller outlined in Section~\ref{sec:control} in the simulator to transform the robot from UGV mode into the upright MIP mode and drive the robot to climb an inclined surface. Figure~\ref{fig:plot_states}-A shows the composite image of the simulation where the robot transformed and drove on a 15 degrees inclined surface, while Fig.~\ref{fig:plot_states}-B shows the states, state references, and the controller inputs into the system. 

We performed the simulation with a controller time step of 0.005 second, an MPC horizon window of 10 steps, and the following weighting parameters: $\vect{Q} = \mathrm{diag}([10,10000,5,25])$ and $\vect{R} = \mathrm{diag}([5,0.5])$. The optimizer in \eqref{eq:mpc_formulation} was solved using the sequential quadratic programming (SQP) algorithm. The heavy weighting on the body angle ($\theta$) is crucial to keep the inverted pendulum dynamically stable but prevents the controller from effectively tracking the wheel speed ($\dot \phi$), as shown in Fig.~\ref{fig:plot_states}-B. Figure~\ref{fig:plot_mpc} shows the MPC cost function ($J$) and the constraint equations of the MPC ($\vect h_c$) using either the ground normal force in \eqref{eq:ground_friction} or the simulator's applied forces from the Simcape's multibody contact forces library. The values of $\vect{h}_c$ are similar between the ground normal force model and the Simscape which shows the accuracy of the model derived in \eqref{eq:ground_friction}. However, the cost $J$ is steadily going up which can be attributed to the poor wheel position and velocity tracking as shown in Fig.~\ref{fig:plot_states}-B. This problem could be resolved using a different weighting and horizon window, or a different optimization solver algorithm.

We have performed preliminary experiments using our robot to perform the UGV to MIP transformation and vice versa, thruster-assisted self-uprighting, landing sequences, and the WAIR wall climbing maneuver. The UGV/MIP transformation sequences can be seen in Fig.~\ref{fig:m4_experiment}, where the self-uprighting and landing sequences are done smoothly using the thrusters. Once the robot is fully upright, the robot can be stabilized using a standard mobile inverted pendulum controller. The WAIR experiment can be seen in Fig.~\ref{fig:wair_experiment}, where the robot in MIP mode climbs a 45 degrees inclined surface by using both the wheels and thruster forces. This slope is steep enough that the robot is unable to climb it in UGV mode. These experiments show how the thrusters and leg transformation can enable the robot to perform tasks that can't be done in UGV form or using just the wheel actuators. 



\section{Conclusions and Future Work}

We have presented the dynamical modeling and simulations of a 2D thruster-assisted MIP model for driving on an inclined surface and using the thrusters to perform WAIR. The simulation shows that the MPC works well in stabilizing the robot and the accuracy of our dynamic and ground friction models. We also shown the feasibility of the thruster-assisted transformation between UGV and MIP modes, in addition to the WAIR using the thrusters to climb steep inclines through experiments. 

The methodology outlined in this work can potentially be extended into a 3D problem, which is significantly more difficult due to the nonholonomic nature of the MIP system. The 3D implementation of the model and experimentations using our robot will be a part of our future work, in addition to other work related to the unique morphing and multi-modal capabilities of our robot, such as morphing mid-flight to perform bounding flight \cite{rayner1985bounding} and multi-modal autonomous navigation.

\nocite{sihite2022unsteady, ramezani2021generative}







\printbibliography

\end{document}